\documentclass{article}
\usepackage{iclr2025_conference,times}

\usepackage{amsmath,amsfonts,bm}

\def\eqref#1{equation~\ref{#1}}

\def\1{\bm{1}}

\DeclareMathAlphabet{\mathsfit}{\encodingdefault}{\sfdefault}{m}{sl}
\SetMathAlphabet{\mathsfit}{bold}{\encodingdefault}{\sfdefault}{bx}{n}

\input{config.sty}
\input{glossary.sty}

\title{Hierarchically Encapsulated Representation for Protocol Design in Self-Driving Labs}

\iclrfinalcopy 

\author{
Yu-Zhe Shi$^{1\,\star{}}$, Mingchen Liu$^{2\,\star{}}$, Fanxu Meng$^1$, Qiao Xu$^1$, Zhangqian Bi$^2$, Kun He$^2$,\\ \ \textbf{Lecheng Ruan}$^{1\, \textrm{\Letter}}$, \textbf{Qining Wang}$^{1\, \textrm{\Letter}}$\\
$^1$ Department of Advanced Manufacturing and Robotics, Peking University\\
$^2$ School of Computer Science and Technology, Huazhong University of Science and Technology\\
$^\star{}$Equal contribution\quad
\Letter\phantom\,\,\texttt{ruanlecheng@ucla.edu, qiningwang@pku.edu.cn}
\vspace{-9pt}
}

\begin{document}

\maketitle

\begin{abstract}
    Self-driving laboratories have begun to replace human experimenters in performing single experimental skills or predetermined experimental protocols. However, as the pace of idea iteration in scientific research has been intensified by Artificial Intelligence, the demand for rapid design of new protocols for new discoveries become evident. Efforts to automate protocol design have been initiated, but the capabilities of knowledge-based machine designers, such as Large Language Models, have not been fully elicited, probably for the absence of a systematic representation of experimental knowledge, as opposed to isolated, flatten pieces of information. To tackle this issue, we propose a multi-faceted, multi-scale representation, where instance actions, generalized operations, and product flow models are hierarchically encapsulated using Domain-Specific Languages. We further develop a data-driven algorithm based on non-parametric modeling that autonomously customizes these representations for specific domains. The proposed representation is equipped with various machine designers to manage protocol design tasks, including planning, modification, and adjustment. The results demonstrate that the proposed method could effectively complement Large Language Models in the protocol design process, serving as an auxiliary module in the realm of machine-assisted scientific exploration.
\end{abstract}

\section{Introduction}\label{sec:intro}

The rapid advancement of \ac{ai} models for the assistance of scientific discovery~\citep{wang2023scientific} has precipitated an increased demand for rapid iteration of ideas, from the generation to the verification of hypotheses. Although \ac{ai} models have expedited the process of hypothesis generation, the validation phase still requires intensive empirical experimentation from human. The concept of self-driving laboratory has been introduced to substantially accelerate the validation process, in organic chemical synthesis~\citep{mehr2020universal,burger2020mobile}, cell biology for medical research~\citep{kanda2022robotic}, and novel material discovery~\citep{szymanski2023autonomous}. With the expertise and effort of experimental scientists and automation engineers, mobile robots and \ac{iot} pipelines are configured to perform a sequence of actions in accordance with a detailed description of the specific experimental procedure, referred to as the \emph{protocol}. 

While existing protocols suffice for some experimental tasks, discovery processes often demand a higher degree of specificity, including: (i) confirmation of unverified experimental objectives to seek specific findings; (ii) testing parallel hypotheses or solutions; and (iii) replication of established experiments within the constraints of available laboratory resources. These necessitate the \emph{design} of new protocols, going beyond the reuse of existing ones available in the protocol databases. Particularly, this includes the \emph{planning} of novel protocols, and the \emph{modification} and \emph{adjustment} of current protocols as appropriate, respectively. Unfortunately, self-driving laboratories currently only execute isolated and duplicated experimental skills~\citep{bedard2018reconfigurable,steiner2019organic}, or pre-specified protocols with sequential actions~\citep{rohrbach2022digitization,manzano2022autonomous}. Any innovation in protocols imposes intensive manual design burden~\citep{mcnutt2014reproducibility,baker20161}, potentially becoming a bottleneck in accelerating scientific discovery. Consequently, there is a quest for the automatic design of protocols tailored to specific goals for self-driving laboratories.

Designing new protocols is a non-trivial task even for human scientists. Novice scientists tend to adhere strictly to established protocols and may be at a loss when faced with the need for variations, from minor adjustments like different available devices to more significant shifts in the overall experimental goal. In contrast, veteran scientists typically have the capability to create or modify protocols as needed, from variations in available resources (\emph{``what I have''}) to desired outcomes (\emph{``what I want''}), even in situations where a similar protocol was not encountered before.

\begin{figure}[t]
    \centering
    \includegraphics[width=\linewidth]{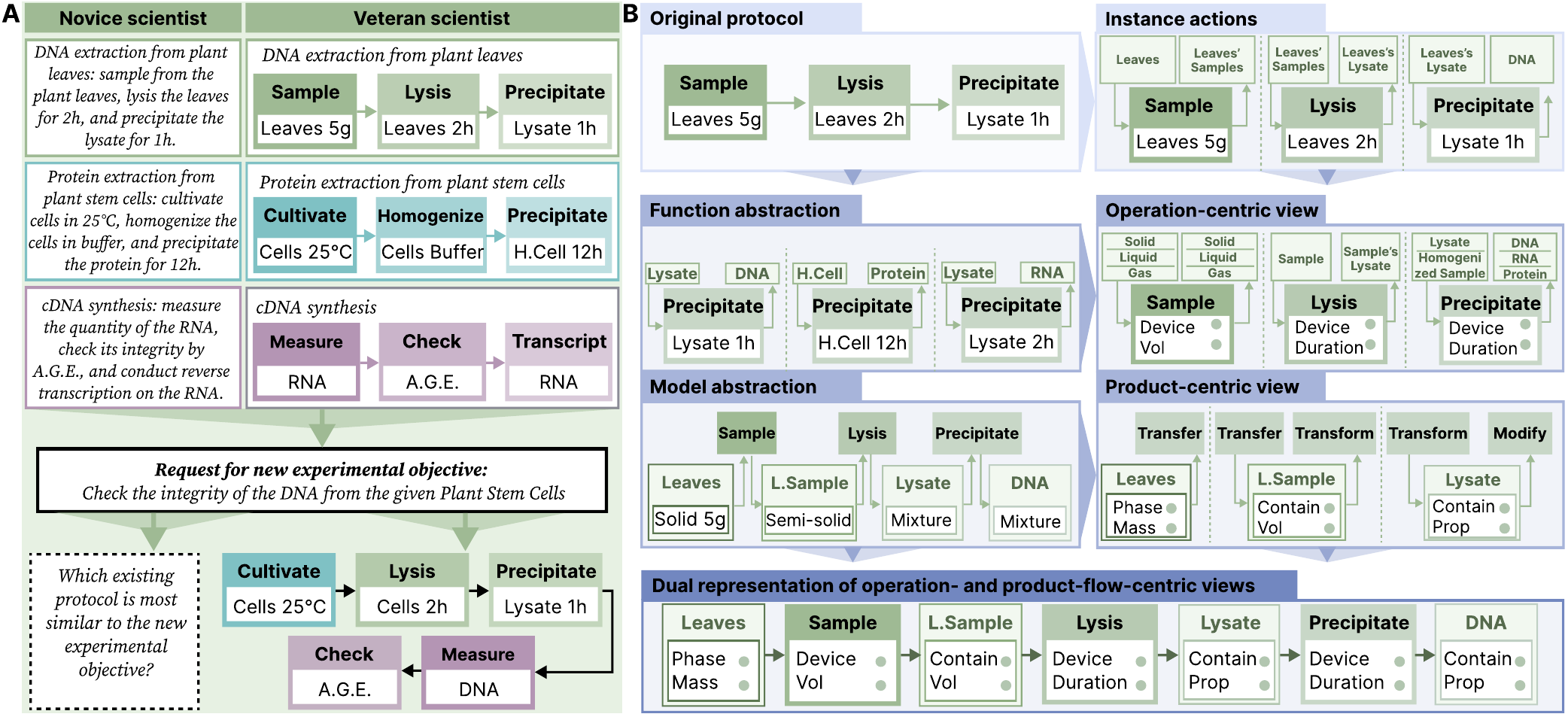}
    \vspace{-2\baselineskip}
    \caption{\textbf{The representations for protocol design.} \textbf{(A)} The example of protocol design by novice and veteran experimental scientists. \textbf{(B)} The hierarchies of our proposed representation, from original full protocol representation, to dual representation of operation- and product-flow-centric views.}
    \label{fig:representation}
    \vspace{-\baselineskip}
\end{figure}

The distinction arises because veteran scientists possess a \emph{systematic} understanding of every ingredient and procedure, contextualizing them globally within the domain of experiment. They know \emph{``what kind of ingredient is used for what purposes''} and \emph{``what kind of operation is used under what conditions'}, while novice scientists mechanically memorize the sequential execution orders and corresponding parameters in a local context. This systematic understanding, or \emph{conceptual knowledge}~\citep{ryle1949concept}, includes the background knowledge of ingredients and atomic operations, as well as the relationships between them. Experienced experimental scientists develop such conceptual knowledge as a \emph{representation} for protocol design~\citep{mccarthy1959programs}, which serves as the vehicle for reasoning processes. Reasoning over conceptual knowledge leverages the rich context of generalized, abstracted concepts of ingredients and operations rather than specified, instantialized ones, which spans a semantic space where originally isolated dots are connected with each other, thereby enhancing the simplicity and flexibility of protocol design~\citep{boden1980artificial,newell1982knowledge}. In summary, veteran scientists' capability to design new protocols stems from an appropriate representation of background knowledge that supports reasoning processes (see \cref{fig:representation}A). 

To implement automatic protocol design on machines, a reasonable choice may be leveraging a \ac{llm}. Trained on extensive corpora, including scientific documents, \acp{llm} possess the potential to facilitate protocol design with the corresponding background knowledge~\citep{ai4science2023impact}. Recently, researchers have made beneficial attempts to design new protocols using \acp{llm} based on descriptions of new experimental goals~\citep{boiko2023autonomous,m2024augmenting}. Regrettably, benchmarking results indicate that the expected capability of \acp{llm} in protocol design is not fully elicited~\citep{o2023bioplanner}. One significant limitation is that \acp{llm} excel at generating new protocols similar to existing ones, \ie, protocols with similar sequential execution orders, but fail to generate those with distinct dependency distributions. This limitation hampers \acp{llm} in scenarios where experimental goals change in high intensity. Another limitation is that the generated protocols sometimes lose critical configuration details for operation execution, necessitating manual correction. These empirical evidences suggest that \acp{llm} exhibit limitations akin to those of novice human experts, implying that \acp{llm} may necessitate a more suitable representation of background knowledge to fully unleash their potential in protocol design. 

Protocol design is a multi-faceted, multi-scale effort requiring the integration of information from different perspectives, from low-level to high-level. This information includes detailed configurations of each atomic operation, temporal relationships between atomic operations, the scope of application for atomic operations with the same reference name, and the reactive relationships between reagents and operations. While \acp{llm} undoubtedly capture such knowledge from their training corpora, the pieces of knowledge remain isolated, unorganized, and not articulated. These flatten background knowledge, rather than conceptual knowledge, hinders \acp{llm} from \emph{flying over} a global view of the novel objectives and \emph{diving into} the details of operations. Therefore, we propose developing a \textbf{multi-faceted and multi-scale representation} for protocol design that provides the designer, such as \acp{llm}, with a vehicle to reason over conceptual knowledge of ingredients and procedures.

We draw inspiration from both cognitive science literature on rationality~\citep{monsell2003task,griffiths2020understanding}, which suggests that we cannot consider information from different views and scales in a single thread~\citep{shi2023perslearn}. We also learn from computer science literature on hierarchical abstraction~\citep{liskov1987keynote}, which indicates that higher-level abstraction semantics possess more powerful expressivity compared to their lower-level counterparts~\citep{abelson1996structure,hopcroft1996introduction}. Combining these insights, we suggest that our desired representation should encapsulate information of different granularities in corresponding hierarchies of abstraction, gaining global design insights with higher-level semantics while completing execution configurations with lower-level semantics. Specifically, we investigate three levels of encapsulation (see \cref{fig:result}B). Starting from the set of original protocols, namely the basic level, we have (i) \textbf{protocol element instantialization}, which decomposes full protocols into instance operations with attributes, within the local context of the specific protocol, resulting a structural representation of the elementary information; (ii) \textbf{function abstraction}, which offers an operation-centric view that generalizes the precondition, postcondition, and execution configurations of each operation in the global context of the experiment domain, resulting a sequential representation of the operations; (iii) \textbf{model abstraction}, which offers an reagent and intermediate product centric view that unifies the status transitions in the global context of the experiment domain, resulting a continuous representation of the experimental environment. This hierarchical structure provides the designer with a representation to consider all possible associations among operations, among products, and between operations and products, with a high degree of freedom, by disentangling originally intertwined information. We implement the representation using \acp{dsl}. The hierarchical syntax of \acp{dsl} maintains both the abstract semantics at the high-level and the precise information at the low-level. Furthermore, the \emph{compositionality} of \ac{dsl} syntax facilitates the flexible protocol designs, addressing the ``flying over global views'' requirement; while \ac{dsl} program verification over the generated protocols upholds their \emph{soundness} and \emph{completeness}; addressing the ``diving into details'' requirements. 

However, the proposed representation does not come without drawbacks---it can be highly dependent on domain-specific knowledge~\citep{mernik2005and,fowler2010domain}. The distributions of reagents, operations, and execution dependencies vary significantly across different domains in experimental sciences, such as \emph{Genetics}, \emph{Medical}, \emph{Bioengineering}, and \emph{Ecology}. Manually crafting \acp{dsl} specialized for these domains requires deep integration between domain experts and programming language experts, which is labour-intensive, case-by-case, and costly~\citep{shi2024autodsl,shi2024constraint}. This obstacle hinders the application of our representation to a broader set of domains~\citep{shi2024abstract}. To make the representation specification more affordable, we develop an algorithm that conducts multi-hierarchy encapsulation automatically driven by the domain-specific corpus of existing protocols. Ultimately, we may be able to take a critical step toward closing the loop of \emph{autonomous} scientific discovery by establishing these two building blocks: (i) the automatic generation of \textbf{representation} for protocol design; and (ii) the automatic \textbf{designer} working on the representation. 

Our contributions in this work are three-fold: (i) we identify the problem of representation for protocol design and develop a hierarchically encapsulated representation for protocol design (\cref{sec:represent}); (ii) we propose a data-driven algorithm that automatically generates the representation for protocol design specialized for the domain of application (\cref{sec:automate}); and (iii) we demonstrate the utility of the resulting representation by conducting protocol planning, modification and adjustment tasks using a variety of machine designers across different domains (\cref{sec:result}). This further indicates that our proposed automatic representation generation approach possesses the potential to function as an auxiliary module for \acp{llm}, enhancing their capability on protocol design. 

\section{Representation for protocol design}\label{sec:represent}

In this section, we describe our representation for protocol design (see \cref{fig:representation}B). We first formulate the basic protocol design problem in \cref{subsec:represent-problem}. Afterwards, starting from the original full protocol, we introduce the three hierarchies of representations: (i) structural representation, \ie, instance actions with attributes (\cref{subsec:represent-instance}); (ii) sequential operation-centric representation, \ie, function abstraction (\cref{subsec:represent-operation}); and (iii) continuous product-flow-centric representation, \ie, model abstraction (\cref{subsec:represent-product}). Furthermore, we describe how the dual representation of operation-centric and product-flow-centric views reciprocatively facilitates the verification of the designed protocols in \cref{subsec:represent-verification}.

\subsection{The protocol design problem}\label{subsec:represent-problem}

Protocol design problem $\text{PD}=(\Phi\mid\omega^*,\mathcal{P},\Omega)$ is generating a desired protocol $\Phi$ given the \textbf{new coming} experimental objective $\rho$, domain of experiment $\mathcal{P}$, and available reagents $\Omega$. A protocol $\Phi=\langle\varphi_1,\varphi_2,\dots\rangle$ is a sequence of experimental steps $\varphi_t$. An experimental objective $\omega^*$ is the expected final product of the experiment. Experimental objectives can range from preparing a desired product, to testing the significance of a specific hypothesis and detecting a predicted behavior, with the latter two potentially followed by additional standalone steps for property test, observation, and interpretation. We denote domains of experiment as $\mathcal{P}$, which influences the distributions of protocols by means of the distributions of operations, reagents, and execution orders, \etc. The set of available reagents $\Omega$ includes originally accessible reagents and excludes those requiring production. 

\subsection{Instance actions with attributes}\label{subsec:represent-instance}

Protocols are originally represented in \ac{nl}, which is the representation suitable for humans' comprehension, but not for machines~\citep{bartley2023building}. Without a syntax decomposing a \ac{nl}-based protocol into information elements precisely, machines are likely to capture only overall, coarse-grained information of protocols and may only retrieve within existing protocols for the one that is most similar to the new experimental objective. Consequently, according to the standards and conventions of experimental sciences~\citep{baker2021five}, the prerequisite of representation for a machine protocol designer should be a structural representation which decompose \ac{nl}-based protocols into instance actions with attributes $\{\varphi_t \mid (\varphi_t^{\text{prec}},\varphi_t^{\text{post}}, \varphi_t^{\text{exec}})\}$. The instance actions are decomposed by execution order and their attributes are the exact context for their execution, namely the precondition $\varphi_t^{\text{prec}}$, \ie, the availability of resources required for this action, postcondition $\varphi_t^{\text{post}}$, \ie, resulting product of the operation, and execution configurations $\varphi_t^{\text{exec}}$. Execution configurations includes the configuration parameters and their corresponding values, \eg, the device for conducting the operation and required experimental conditions such as duration, acidity, and lightening. An instance action can be reusable in another protocol once the execution context is matched. 

With such reusability, we are on the first time to have \emph{building blocks} for constructing a new protocol rather than retrieving existing ones. These building blocks capture fine-grained execution configuration parameters through maintaining the nested data structures of key-value pairs. This structural representation serves as a syntactic constraint on the preciseness of designed protocols. Practical attempts have been made echoing this idea~\citep{o2023bioplanner,leonov2024integrated}. 

\subsection{Operation-centric view with function abstraction}\label{subsec:represent-operation}

The reusability of instance actions with attributes is highly limited, as their semantics are highly specified in the low-level. The total amount of the instance actions can be extremely high, \ie, about 150K per domain, thus the probability of the exact matching between execution contexts can be extremely low. Consider the three different instance actions with attributes \emph{``Homogenization of mouse liver tissue using a bead mill''}, \emph{``Homogenization of bacterial cell suspension using an ultrasonic homogenizer''}, and \emph{``Homogenization of bacterial air samples using a nebulizer''}. Although they come with totally different preconditions, postconditions, and execution configurations, particularly the required device varying according to the phase of the experimental subject, they share the semantic identifier \emph{``Homogenization''} for reference. Sharing semantic identifier indicates that these instance actions share the same \emph{purpose} on the semantics level. In experimental sciences, \emph{``Homogenization''} always refers to the breakdown of a sample into a uniform mixture. Whether it's tissue, cell suspension, or gas doesn't change the purpose of the operation. This is critical for protocol design, since it essentially requires satisfying the ultimate goal through a series of subgoals. Therefore, the desired representation should generalize the semantics of operations to any possible contexts in the corresponding domain of experiment, rather than only specific contexts. 

We implement such generalization by encapsulating varied instances of preconditions, postconditions, and execution configurations into an \emph{interface} for the operation. Namely, we refer to an operation with semantic identifier $\varphi$ through an interface $\phi$ to a set of execution contexts, in the form of $\langle\varphi\mapsto\phi\mapsto\{(\varphi^{\text{prec}},\varphi^{\text{post}}, \varphi^{\text{exec}})\}\rangle$. The operation $\varphi$ can be grounded to a corresponding instance action in any matched execution contexts, echoing \emph{modular design}~\citep{hirtz2002functional}. The reusability of encapsulated operations comes with greater significance than that of instance actions, as there are only about 1K operations per domain in total, which is only $1/150$ of that of instance actions. As flexible building blocks, operations can be easily fitted into any breakpoints with suitable preconditions and postconditions in the constructing experiment sequence. This sequential representation of the operations serves as a semantic constraint on the compact permissible set of primitives for protocol design~\citep{shi2023complexity}, maintaining both degree of freedom and correctness.

\subsection{Product-flow-centric view with model abstraction}\label{subsec:represent-product}

Sequence of operations make up of protocols. However, operations are the methods to realize rather than the objectives to achieve. For experimental objectives of testing, preparing, or detecting~\citep{schwab2020different}, the common focus is always the specific status of final product, not the operations. Starting from initial reagents, the status of product flow is manipulated step-by-step by the operations, till the final product. Unfortunately, the information of product status transition is \emph{latent} in protocols and is \emph{twisted with} descriptions of experimental steps. For the operation-centric view, the transitions of product flow statuses remains a \emph{black box environment}. For example, the operation description \emph{``Centrifuge the tubes at 15,000 x g for 20 minutes''} does not directly reveal the transition from product in mixture status to products in distinct phases. The lack of coherent tracking of the product flow is problematic of protocol design, as the product flow holds \emph{spatial-temporal invariance}, just the same as the general physical environment. Status transitions of the product flow are primarily caused by the effects of operations, thereby it serves as the \emph{invariant} in executing the protocol from the perspective of programming. Therefore, the desired representation should also serve as the \emph{model} interacting with the sequence of operations. 

To disentangle product status from their latent representation in the operation-centric view, we propose an explicit product flow centric view that tracks the status of the product flows with detail, such as component, volume, container, and other physical and chemical properties of the product, and also the predecessor operation that yields the product and the successor operation that takes the product as input. Each product flow unit, \ie, one individual component in the product flow between two adjacent steps, is an instance with attributes $\{\omega_t\mid(\omega_t^{\text{pred}},\omega_t^{\text{succ}},\omega_t^{\text{prop}})\}$. Analogous to the generalization of operations' semantics, product flow units share commonalities between components with the same semantic identifier for reference---they may share a specific range of predecessor operations $\omega^{\text{pred}}$ and successor operations $\omega^{\text{succ}}$, and a selected set of key properties to consider $\omega^{\text{prop}}$. For example, the \emph{``supernatant''} is usually generated by a \emph{``centrifugation''} operation, passing into \emph{``filtration''} or \emph{``spectrophotometric analysis''}, and focusing on the properties \emph{acidity} and \emph{viscosity} rather than other possible properties. Thus, we encapsulate the information of contexts and properties into the semantics of product flow units, in the form of $\langle\omega\mapsto(\omega^{\text{pred}},\omega^{\text{succ}},\omega^{\text{prop}})\rangle$. As solid pipelines bridging the building blocks, product flow units can verify the coherency of the entire designed protocol. This continuous representation of the environments serves as a program verifier, checking the prerequisite and simulating the effect of each operation, alleviating unpredictable behaviors among the interaction between operations and product flows.

\subsection{Reciprocative verification over the dual representation}\label{subsec:represent-verification}

\input{algorithms/reciprocative_verification.sty}

The dual representation of operation-centric and product-flow-centric views intrinsically equips with a verification mechanism through a reciprocative process akin to two interacting threads. The first thread focuses on verifying the operation flow, taking as input an operation $\varphi_t$ along with its precondition $\varphi_t^{\text{prec}}$ and postcondition $\varphi_t^{\text{post}}$. The second thread handles the verification of the product flow, taking as input a product $\omega_t$ along with its predecessor operation $\omega_t^{\text{pred}}$ and successor operation $\omega_t^{\text{succ}}$.

Specifically, for the operation verification (corresponding to \textsc{OFVerification} in \cref{alg:verification}), we ensure that each operation can be correctly executed given its input reagents and that it yields the expected output products. This involves checking that the preconditions are satisfied by the available products from preceding operations and that the postconditions are well-defined for subsequent use. Concurrently, the product flow verification (corresponding to \textsc{PFVerification} in \cref{alg:verification}) involves tracking each unit of product flow through the protocol. We verify that the product is generated by the specified operation and that it possesses the necessary properties $\omega_t^{\text{prop}}$ for consumption by the next operation. 

The interaction between these two threads forms a feedback loop where the verification of operations and products mutually inform and constrain each other. This reciprocative method allows us to iteratively refine the protocol, ensuring that each step is both operationally feasible and chemically coherent. \acp{llm} are employed to implement the functions \textsc{CheckOpConditions} and \textsc{CheckProperties}, extracting and verifying operation conditions and product properties from natural language protocol descriptions through instruction-following in-context learning \citep{wei2021finetuned,brown2020language}. For the prompts employed, readers are referred to \cref{subsec:supp-implement-external}.

\begin{figure}[t]
    \centering
    \includegraphics[width=\linewidth]{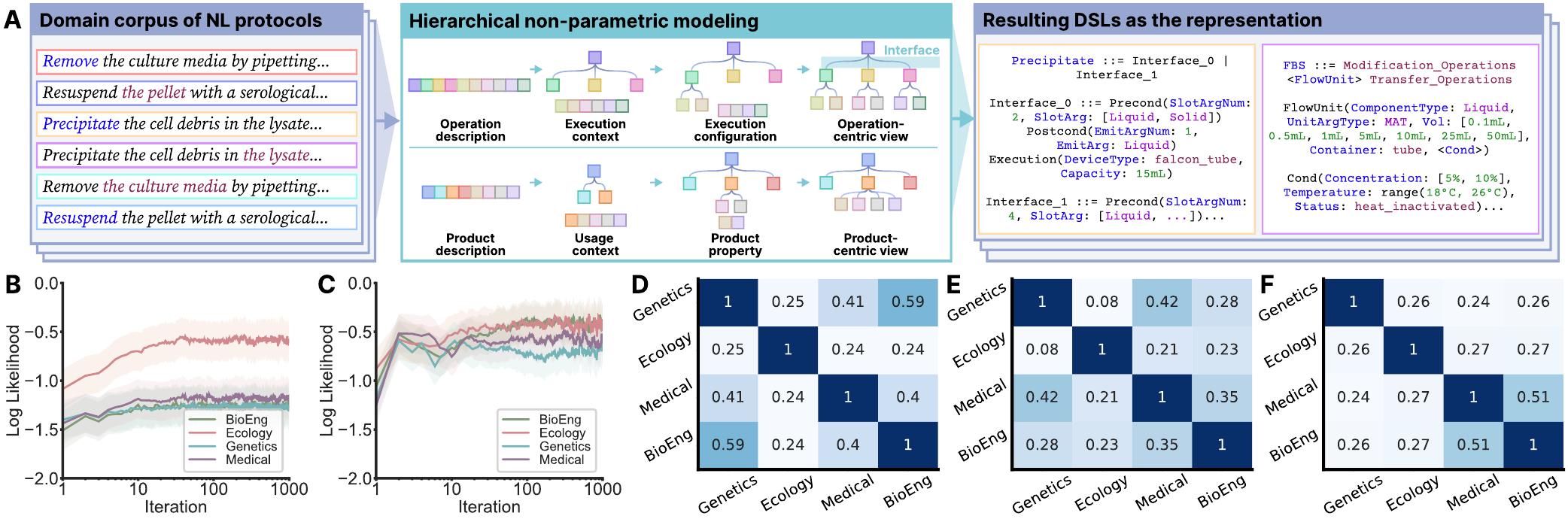}
    \vspace{-2\baselineskip}
    \caption{\textbf{Diagram of automatic representation generation.} \textbf{(A)} Illustration of the workflow. \textbf{(B)} Convergence curve of automatic function abstraction. \textbf{(C)} Convergence curve of automatic model abstraction. \textbf{(D-F)} Confusion matrices on operation distribution (D), product distribution (E), and device distribution (F), between \acp{dsl} across domains. Correlation scores are low except the ones along the diagonals, indicating the significant inter-domain distinctions between the resulting \acp{dsl}.}
    \label{fig:automate}
    \vspace{-\baselineskip}
\end{figure}

\section{Automatic representation generation}\label{sec:automate}

In this section, we describe the proposed data-driven algorithm to automatically generate the hierarchically encapsulated representation for protocol design (see \cref{fig:automate}A). We first define the problem of generating the desired representation by means of \ac{dsl} design (\cref{subsec:automate-problem}). We then introduce methods for generating operation-centric (\cref{subsec:automate-function}) and product-flow-centric (\cref{subsec:automate-model}) \ac{dsl} views.

\subsection{The representation generation problem}\label{subsec:automate-problem}

We denote the problem of generating the representation for protocol design within a given domain as $\text{RG}=(\{\langle\varphi\rangle,\langle\omega\rangle\}\mid\mathcal{P},\mathcal{C})$. The representation is a \ac{dsl} with language features accommodating both the operation-centric program view $\langle\varphi\rangle$ and the product-flow-centric program view $\langle\omega\rangle$. The domain-specific corpus $\mathcal{C}=\{\Phi_1,\Phi_2,\dots,\Phi_{|\mathcal{C}|}\}$ consists of \textbf{existing} protocols published in top-quality journals within the corresponding experimental domain. The source and profiles of $\mathcal{C}$ of each domain is detailed in \cref{subsec:supp-dataset-corpus}. We can obtain instance action with attributes based on $\mathcal{C}$ in a straightforward way through \ac{nl} information extraction (see \cref{subsec:supp-implement-preprocess} for implementation details). The prior knowledge of operations and products, $p(\varphi)$ and $p(\omega)$, including the basic syntax of the key-value structures and the elementary taxonomies, is derived according to the general commonsense of experimental sciences, as aforementioned in \cref{sec:represent}. Specifically, the problem essentially aims to fit the joint distribution models $p(\varphi,\phi, \varphi^{\text{prec}},\varphi^{\text{post}}, \varphi^{\text{exec}})$ and $p(\omega,\omega^{\text{pred}},\omega^{\text{succ}},\omega^{\text{prop}})$ with domain-specific corpus $\mathcal{C}$ given prior knowledge $p(\varphi)$ and $p(\omega)$.

\subsection{Automatic function abstraction}\label{subsec:automate-function}

The key challenge of encapsulating the operation-centric view is to aggregate all possible execution contexts for an operation, and then generalize the contexts to the interface. If we keep each of the use case as one single instance of the interface, which can be in thousands regarding one operation, the generalization is meaningless. Since there is no prior knowledge about the interface in advance, we develop the algorithm following the idea of non-parametric modeling, \ie, \ac{dpmm}, resulting in flexible identification of interface instances.  

\paragraph{Hierarchical non-parametric modeling}

As we must handle information coming in different granularities, from interface structures to values of parameters, we choose to model the operations in a hierarchical fashion. Compared with the flatten spectral clustering approach developed by~\citet{shi2024autodsl}, which compresses all information of an operation into a embedding vector, our modeling is competent for considering information at different levels comprehensively. We carefully adopt the prerequisite that the interface is generated subject to the operation, preconditions, postconditions, and execution configurations are generated subject to the interface, and the value of configuration parameters are generated subject to their corresponding keys. Thus, we have the model:

\begin{equation}
    \begin{aligned}
        &p(\varphi,\phi, \varphi^{\text{prec}},\varphi^{\text{post}}, \varphi^{\text{exec}}, \varphi^{\text{exec-v}})\\&=p(\varphi^{\text{exec-v}}\mid\varphi,\phi,\varphi^{\text{exec}})p(\varphi^{\text{exec}}\mid\varphi,\phi)p(\varphi^{\text{prec}}\mid\varphi,\phi)p(\varphi^{\text{post}}\mid\varphi,\phi)p(\phi\mid\varphi)p(\varphi),
    \end{aligned}
\end{equation}

where $\varphi^{\text{exec-v}}$ denotes the values of configuration parameters. Within each iteration of the \ac{dpmm} process, we sample the variables level-by-level. Since the structures of preconditions, postconditions, and the selection of devices and configuration parameters are discrete, we sample them directly from the Dirichlet Process. As permissible values of parameters can be discrete, \eg, an array of specific values, common in acidity preparation; continuous, \eg, an interval with minimum and maximum values, common in temperature setting; or mixed, \eg, an array of specific values with random perturbations around the mean, common in timing control, we conduct the sampling by integrating Gaussian Process with Dirichlet Process $\varphi^{\text{exec-v}}\mid\varphi,\phi, \varphi^{\text{exec}}\sim DP(\alpha,H(\varphi^{\text{exec}}), \phi, \varphi)\times GP(m,K)$, where $\alpha$, $H$, $m$, and $K$ are corresponding hyperparameters. 

\paragraph{Unification of the interface}

While clustering similar interface instances encapsulates operations, there may remain redundant interfaces due to minor discrepancies. These discrepancies often arise from differences in parameter values or naming conventions that do not fundamentally alter the operation's functionality. To alleviate such redundancies, we implement a unification process for the interfaces. Specifically, interface instances associated with the same operation are considered equivalent if they have the same number of slots and emits and share the same keys in their execution configuration parameters. By abstracting away differences in parameter values and names, we unify these interfaces into a single, generalized interface, akin to the algorithm proposed by~\citet{martelli1982efficient}. Unification enhances the generality of the operation-centric view by consolidating functionally-identical interfaces, maintaining a concise and representative set of operations.

\begin{figure}[t]
    \centering
    \includegraphics[width=\linewidth]{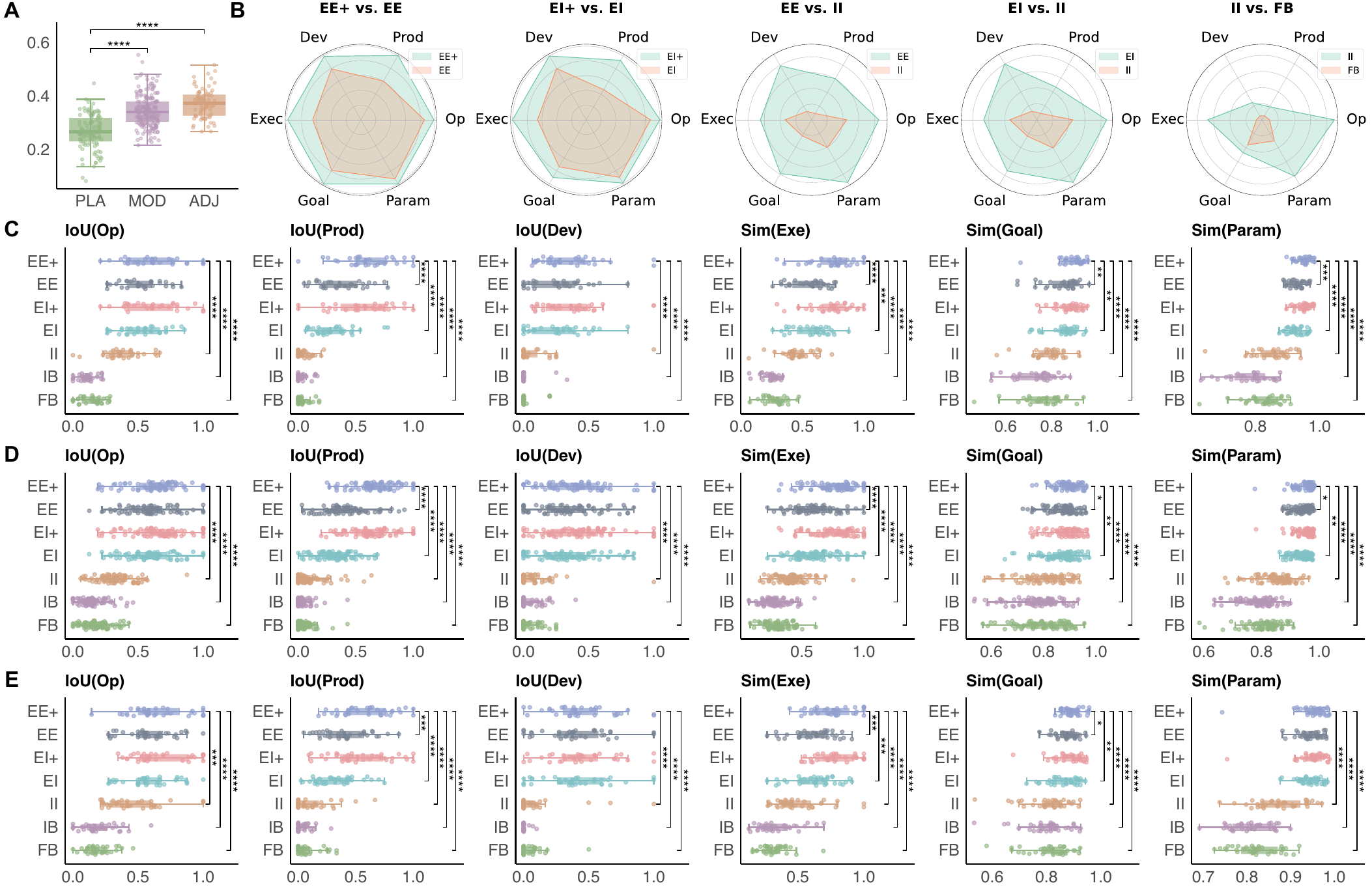}
    \vspace{-2\baselineskip}
    \caption{\textbf{Results of protocol design.} \textbf{(A)} Profile of text-level similarity between testing sets of the three tasks. \textbf{(B)} Pairwise comparison between the capabilities of different machine designers across the six dimensions. \textbf{(C-E)} Performances of the seven machine designers on the planning (C), modification (D), and adjustment (E) tasks across the six dimensions (index by column).}
    \vspace{-\baselineskip}
    \label{fig:result}
\end{figure}

\paragraph{Results}

Function abstraction converges on the domains respectively, as shown by the likelihood curve yielded by non-parametric model in \cref{fig:automate}B. In the \ac{dsl} of Genetics, there are $304$ operations in total, with an average of $7.9$ interface instances per operation; for Medical, these two quantities are $269$ and $6.9$; for Bioengineering, they are $196$ and $7.8$; and for Ecology, they are $100$ and $3.5$. We find that a majority of operations with high occurrence frequency are unique to one domain, such as \texttt{Pipette} to Medical and \texttt{Lyse} to Genetics (see \cref{fig:automate}D). There are also common operations across domains, such as \texttt{Concentrate} and \texttt{Culture}. Take \texttt{Concentrate} for an example, its interface captures the instances with different devices according to input phases, \eg, use \texttt{Bench-top\_centrifuge} for \texttt{Liquid} while \texttt{Isotope\_separation\_centrifuge} for \texttt{Gas}, and also instances with different emits, \eg, selecting \texttt{Supernatant} or \texttt{Suspension} as the product to keep.

\subsection{Automatic model abstraction}\label{subsec:automate-model}

The key challenge of encapsulating the product-flow-centric view is to select proper descriptive properties of a flow unit component. There exists false positive cases, where properties are attributed to components with the same semantic identifier but in different phases, \eg, we consider ethanol with the property volume when it comes in liquid and with the property pressure when it comes in gas. There also exists false negative cases, where exact same components are regarded as different ones due to different reference names, \eg, Acetylsalicylic Acid, ASA, and Aspirin refer to the same thing. To alleviate false positive and false negative results, we discard the design choice of the interface in the operation-centric view, which tends to cover the possibly richest context, and thereby have the non-parametric model:

\begin{equation}
    \begin{aligned}
        &p(\omega,\omega^{\text{pred}},\omega^{\text{succ}},\omega^{\text{prop}},\omega^{\text{prop-v}})\\&=p(\omega^{\text{prop-v}}\mid\omega^{\text{prop}},\omega)p(\omega^{\text{prop}}\mid\omega)p(\omega^{\text{pred}}\mid\omega)p(\omega^{\text{succ}}\mid\omega)p(\omega),
    \end{aligned}
\end{equation}

where $\omega^{\text{prop-v}}$ denotes the values of property parameters.

\paragraph{Results}

Model abstraction converges on the domains respectively, as shown by the likelihood curve in \cref{fig:automate}C. In the \ac{dsl} of Genetics, there are $17,190$ model states, \ie, product flow unit as product status, in total; the quantity is $12,472$ for Medical; $11,418$ for Bioengineering; and $2,205$ for Ecology. We find that most components of product flow units with high occurrence frequency are unique to one domain, such as \texttt{RNA} to Genetics and \texttt{HCC} to Medical (see \cref{fig:automate}E/F). Take \texttt{Ethanol} for example, the model captures its possible concentrations in liquid rather than in gas.

\section{Experiments and discussion}\label{sec:result}

In this section, we report and discuss the results of our experiments. We start from describing our realistic novel protocol design tasks (\cref{subsec:result-task}), along with the metrics to measure the consistency between the designed protocol and the groundtruth protocol (\cref{subsec:result-metrics}). Afterwards, we introduce the alternative representations and machine designers used for comparison (\cref{subsec:result-designers}). Finally, we report and analyze the experimental results both quantitatively and qualitatively (\cref{subsec:result-res}). 

\subsection{Protocol design tasks}\label{subsec:result-task}

\input{tables/dataset_stat.sty}

Generating unverified experimental objectives and their corresponding protocols specially for our protocol design tasks is impractical because those experiments which have not been peer-reviewed and published can be problematic regarding the contents themselves. To maintain both reality and scale of the testing set, for each domain we \textbf{filter out} a small subset of protocols which significantly differ from the remaining major part of the protocol set and \textbf{exclude} this subset from the corpora for automatic representation generation (\cref{subsec:supp-dataset-corpus}). This selected subset form the groundtruth of the testing set.

We exploit quantitative indicators to assist testing set selection, which follows the convention of measuring a protocol's \emph{novelty} in experimental sciences~\citep{schwab2020different}. We comprehensively consider three indicators: (i) similarity between the text embedding of the \ac{nl}-based description of purpose of protocols, employing the evaluation model in~\citet{o2023bioplanner}; (ii) \ac{iou} between the instance actions of protocols; (iii) similarity between the execution sequence of protocols, implemented through the \ac{sa} algorithm~\citep{smith1981identification}. To note, indicators (ii) and (iii) are calculated upon the protocols pre-processed by the workflow described in \cref{subsec:supp-implement-preprocess}. Indicator (i) captures the high-level idea of protocol design, indicator (ii) is correlated to the implementation of the protocol design, while indicator (iii) captures the low-level information of protocol execution.

In response to the three purposes of protocol design introduced in \cref{sec:intro}, we specify the planning, modification, and adjustment tasks of protocol design. Candidate planning tasks, which are the confirmation of unverified experimental goals, come with relatively low scores (within the $20\%$ lowest) on indicators (i) and (ii). Candidate modification tasks come with fair scores (around the $40\%$ lowest) on indicators (i) and (ii) and relative low score on indicator (iii). Candidate adjustment tasks come with relatively high scores (within the $40\%$ highest) on all of the three indicators. 

We obtain the final testing set through a human-machine collaborative workflow. We first detect the outliers of the original protocol corpus of each domain under the metrics above, thereby forming a candidate set. Afterwards, experts of the corresponding domain (holding at least a Master's degree majoring in that domain) manually check the applicability of protocols in the candidate set with cross-validation, discarding the misclassified ones, requesting for more candidate protocols, and refining the groundtruth file when necessary. The testing set includes 140 new protocols and 1757 steps in total, across the domains of Genetics, Medical, Bioengineering, and Ecology, with $23\%$ for planning, $52\%$ for modification, and $25\%$ for adjustment (see \cref{tab:testset-stat} and \cref{fig:result}A for details).

\subsection{Inter-protocol consistency metrics}\label{subsec:result-metrics}

Evaluating the consistency between a designed protocol and the groundtruth is not like comparing between two plain strings~\citep{o2023bioplanner}. Based on the corresponding commonground in experimental sciences~\citep{bartley2023building}, we design six-dimensional metrics to comprehensively cover all of the major factors without biased weighting and composition. The six dimensions include: (i) \textbf{IoU on operations}, $\text{IoU(Op)}=\text{IoU}(\{\varphi_{1\dots|\Phi|}\},\{\varphi'_{1\dots|\Phi'|}\})$, \ac{iou} between instance actions of the designed protocol $\Phi$ and the groundtruth $\Phi'$; (ii) \textbf{IoU on reagents and intermediate products}, $\text{IoU(Prod)}=\text{IoU}(\{\omega_{0\dots|\Phi|}\},\{\omega'_{0\dots|\Phi'|}\})$; (iii) \textbf{IoU on devices}, $\text{IoU(Dev)}=\text{IoU}(\{\varphi(\text{Dev})_{1\dots|\Phi|}\},\{\varphi(\text{Dev})'_{1\dots|\Phi'|}\})$, where $\varphi(\text{Dev})_t$ denotes the exact device for conducting the instance action $\varphi_t$; (iv) \textbf{Similarity between the execution sequences}, $\text{Sim(Exec)}=\text{SeqAlign}(\langle\varphi_{0\dots|\Phi|}\rangle, \langle\varphi'_{0\dots|\Phi'|}\rangle)$, where $\text{SeqAlign}(\cdot,\cdot)$ denotes the ordered sequence similarity score calculation by the \ac{sa} algorithm; (v) \textbf{Similarity between experimental objectives}, $\text{Sim(Goal)}=\text{Cos}(S(\rho),S(\rho'))$, where $S(\cdot)$ represents the serialization operation on structural representations of protocols; (vi) \textbf{Similarity between complete protocols at parameter-wise level}, $\text{Sim(Param)}=\text{Cos}(S(\Phi),S(\Phi'))$. These six dimensions capture protocol information from low to high granularities, and also measure the consistency of both ingredient knowledge and procedural knowledge, offering a relatively objective evaluation standard.

\subsection{Machine designers}\label{subsec:result-designers}

We implement an array of designers by combining different representations with different \ac{llm}-based automatic designers under tractable computing load (see \cref{subsec:supp-implement-load}). We investigate four types of representations, including the original \ac{nl}-based protocol representation (\texttt{Flatten}) and the three levels of encapsulation described in \cref{sec:represent}, \ie, instance actions with attributes (\texttt{Instance}), operation-centric view only (\texttt{Encapsulated}), and the dual representation with operation- and product-flow-centric views (\texttt{Encapsulated+}). We consider three types of \ac{llm}-based protocol designers: (i) \texttt{Baseline}, a pure \ac{llm}-based approach with \ac{rag} on the corresponding corpora (\cref{subsec:supp-implement-baseline}); (ii) \texttt{Internal}, which takes the specific representation as part of the prompt of an \ac{llm}, requesting it to output the protocol under the constraint of the given representation (\cref{subsec:supp-implement-internal}); (iii) \texttt{External}, where the representation serves as an external constraint layer for the output of an \ac{llm}, verifying and refining the designed protocols (\cref{subsec:supp-implement-external}). Notably, the external verifier is part of the resulting \ac{dsl} as our proposed representation for protocol design.

The combination of representation and designer does not span a Cartesian space due to the intrinsic limitations of \texttt{Flatten} and \texttt{Baseline}. Therefore, we implement seven machine designers, including: (i) \texttt{Flatten-Baseline(FB)}, \ac{llm} with \ac{rag} on original protocol corpora; (ii) \texttt{Instance-Baseline(IB)}, \ac{llm} retrieval on the protocol corpora translated into instance actions; (iii) \texttt{Instance-Internal(II)}, prompting \ac{llm} with the \ac{isa} of instance actions, following the implementation of the currently state-of-the-art method~\citet{o2023bioplanner}; (iv) \texttt{Encapsulated-Internal(EI)}, prompting \ac{llm} with the \ac{dsl} with operation-centric view; (v) \texttt{Encapsulated-External(EE)}, \ac{llm} equipping with the external verifier provided by the \ac{dsl} with operation-centric view; (vi) \texttt{Encapsulated-Internal+(EI+)}, prompting \ac{llm} with the \ac{dsl} with the dual representation; and (vii) \texttt{Encapsulated-External+(EE+)}, \ac{llm} equipping with the external verifier provided by the \ac{dsl} with the dual representation.

\subsection{Protocol design results}\label{subsec:result-res}

The complete quantitative results across the four domains, the three tasks, and the six dimensions of evaluation metrics are presented at \cref{sec:supp-result}. Through paired samples t-test, we find that \texttt{EE+} and \texttt{EI+} significantly outperform other alternative approaches (\texttt{EE+} outperforms \texttt{EE}: $t(278)=8.007, \mu_d<0, p<.0001$; \texttt{EI+} outperforms \texttt{EI}: $t(278)=8.397, \mu_d<0, p<.0001$; \texttt{EE+} outperforms \texttt{II}: $t(278)=24.493, \mu_d<0, p<.0001$; \texttt{EI+} outperforms \texttt{II}: $t(278)=23.855, \mu_d<0, p<.0001$; see \cref{fig:result}C-E). These comparisons demonstrate the suitability of our desired representation for protocol design. Similarly, we find that approaches equipping with a relatively higher-level representation significantly outperforms their counterparts with a relatively lower-level representation (\texttt{EE} outperforms \texttt{II}: $t(278)=16.315, \mu_d<0$, $p<.0001$; \texttt{EI} outperforms \texttt{II}: $t(278)=15.259, \mu_d<0$, $p<.0001$; \texttt{II} outperforms \texttt{FB}: $t(278) = 8.340, \mu_d < 0, p < .0001$; see \cref{fig:result}B).

\subsection{Discussion}\label{subsec:result-disc}

This work proposes a hierarchically encapsulated representation for the conceptual knowledge in experimental sciences, including instance actions with attributes, sequential representation of operations with function abstraction, and continuous representation of product-flows with model abstraction, to fully elicit \acp{llm}' capability on protocol design as an auxiliary module. The following discussions on results reveal the design rationality, scalability, and generality of the representation.

\paragraph{Contributions of the building blocks}

The encapsulated representation approaches with dual views outperform their counterparts without dual views by enhancing both intra-step and inter-step details. At the intra-step level, \texttt{EI} and \texttt{EE} offer richer semantic information than \texttt{IB} and \texttt{II}, leveraging protocol-centric view to capture detailed configuration each operation. This feature accounts for their satisfactory performance on \texttt{IoU(Op)}. At the inter-step level, \texttt{EI+} and \texttt{EE+} treat each step as a \texttt{FlowUnit}, incorporating both preceding and succeeding step contexts, leading to notable improvements in \texttt{Sim(Exec)} and \texttt{IoU(Prod)}. This creates a \emph{double assurance} mechanism~\citep{shi2024expert}: the first assurance comes from internal input/output checks within each instruction, and the second from the input/output characteristics inferred from neighboring instructions. Namely, we estimate the output of the preceding operation and check its alignment with the current step's input. This design enhances step linkage, verification, and overall coherence, ensuring higher consistency and robustness in complex protocol workflows. Please refer to \cref{subsec:supp-case-1} for the case study.

\paragraph{Handling different task complexities}

The overall performance aligns with the trend in complexity across the three tasks (\cref{fig:result}A); however, the dual-view encapsulated representations, \texttt{EI+} and \texttt{EE+}, demonstrate superior performance compared to their counterparts. In planning, these methods consider all necessary components, enabling creative yet structured protocol generation. For modification tasks, they provide feedback on parameter changes, detecting inconsistencies that their counterparts might fail to capture. In adjustment tasks, \texttt{EE+}’s external verifier maintains protocol integrity by identifying component relationships. Please refer to \cref{subsec:supp-case-2} for the case study.

\paragraph{Generality across domains}

Our \ac{dsl}-based approaches offer a unified, modular representation with generalizability across scientific domains (see domain-indexed results at \cref{subsec:supp-result-domain}). The dual-view approach abstracts experimental processes into operations and flow units, capturing essential details while remaining applicable across fields. By representing dependencies between steps and tracking product flow, the replication of experiments could be enhanced. The framework captures cross-domain commonalities while allowing domain-specific content like specialized operations and reagents. This unified representation standardizes protocols and enables researchers to adopt experimental protocols from multiple fields, fostering interdisciplinary collaboration and innovation. Please refer to \cref{subsec:supp-case-3} for the case study. Limitations on generality are discussed at \cref{sec:supp-limit}.

\section*{Acknowledgements}

This work was partially supported by the National Natural Science Foundation of China under Grants 52475001. Q. Xu is a visiting student at Peking University from University of Science and Technology of China. The authors would like to thank Haofei Hou for his earlier works regarding domain-specific representations, and also Jiawen Liu for her assistance in figure drawings.

\bibliographystyle{iclr2025_conference}
\bibliography{references}

\clearpage
\newpage
\appendix

\renewcommand\thefigure{A\arabic{figure}}
\setcounter{figure}{0}
\renewcommand\thetable{A\arabic{table}}
\setcounter{table}{0}
\renewcommand\theequation{A\arabic{equation}}
\setcounter{equation}{0}
\pagenumbering{arabic}
\renewcommand*{\thepage}{A\arabic{page}}
\setcounter{footnote}{0}

\section{Additional remarks}\label{sec:supp-remarks}

\subsection{Rationale of the overall design choice}

It seems that we can formulate the protocol design problem in the fashion of \ac{mdp} and solve it by heuristic-based planning methods or \ac{hrl} approaches. However, although the formulation itself is feasible, solving the problem may not be practical. Consider solving the problem through an HRL approach designed for heterogeneous action space with parameters (as the protocol is required to decide both the key properties of an operation and the corresponding values). This hierarchical agent may be trained to converge on a fine-grained environment with a clearly designed reward function, or on a large dataset with trajectories for offline learning. Unfortunately, we have access to neither an interactive environment simulating the experiments nor sufficient data to support offline training~\citep{pateria2021hierarchical}. 

Treating the experimental procedures as a white box and creating digital twins for experiments can be an elegant solution and thereby facilitate various applications other than protocol design. This effort requires elaborated design of simulation granularity, exhaustive collection of primitive principles of the system, efficient implementation of rule production, and define precise metrics for evaluating the distance between current and objective states (serving as a reward function), which can be labor-intensive and is far out of the scope of this work. On the other hand, viewing those published protocols as trajectories for offline training, the scale of the offline dataset and the density of the reward function are much too insufficient to support training to convergence. Augmenting the data, synthesizing realistic trajectories, or enhancing the accessibility of protocols, are out of the scope of this work. Given the current obstacles, we choose not to formulate the problem in an \ac{mdp} fashion. Though an \ac{mdp}-style formulation can be more precise and elegant, it may misguide the readers to some extent. Instead, we decide to leverage the rich domain-specific knowledge provided by knowledge-based agents such as \acp{llm}, where knowledge may complement the lack of data and dense reward function. This design choice is also in line with the initial attempts on automatic experiment design~\citep{boiko2023autonomous,m2024augmenting}. 

In summary, our design choice of formulation is a compromise based on currently limited resources and restricted scope. Nonetheless, the exploration of more precise and elegant formulations represents a promising avenue for future research.

\subsection{Intuition behind the interface}

Interface is a concept of functional abstraction~\citep{abelson1996structure}. Interface disentangles the abstract functionality on the semantics level and its corresponding implementation details on the execution level. This approach encapsulates the implementation of an operation into a \emph{black-box}, so the users of the operation would only need to consider its input and output. Therefore, with such encapsulated representation for protocol design, we only need to care about the consistency between the output of the predecessor operation and the input of the successor operation, without caring about their implementation details. 

This is the idea behind operationalization. Operationalization makes the interface an abstract function over all relative instance actions. The interface is abstracted from the execution contexts of all instance actions with the same reference name, \ie, the same purpose, and can be instantiated to an instance action given a specific execution context. A specific context can be the predecessor operation, the successor operation, the precondition, or the postcondition of the considered operation. An instance action configures a specific implementation for a specific execution context. For the operation \emph{``Homogenization''}, the implementation of one instance action can be \emph{``using an ultrasonic homogenizer''} if the precondition, namely, the execution context, has intermediate product \emph{``cell suspension''} available; the implementation of another instance action can be \emph{``using a bead mill''} if the precondition contains tissue. This example demonstrates the relationship between interface and instance actions of an operation: the interface is abstracted from the set of instance actions and can be instantiated to instance actions.

Here we also give a more intuitive example to enhance the reader's comprehension. Consider the culinary scenario with the actions \emph{``frying the egg''}, \emph{``frying the fish''}, and \emph{``frying the steak''}.These are different instance actions coming with the same purpose \emph{``to fry something''}. Therefore, we can abstract the interface from these instance actions to operationalize the operation \emph{``fry''}. The input of \emph{``fry''} should be something raw and its output should be something fried. Given different preconditions with available eggs or pieces of steak, the abstract semantic operation \emph{``fry''} can be grounded to instance actions \emph{``frying the egg''} or \emph{``frying the steak''} respectively, through the instantiation of the interface. In summary, an interface serves as the bridge between the semantics level and the execution level.

\subsection{Values of manual protocol certification}

Certification is always one of the central focuses in the engineering practices of automation. In our practice, we only automate the process of protocol design, which is the primary objective of this work, and keep the manual certification part. On one hand, relieving experimental scientists from the labour-intensive protocol design tasks, thereby allowing them more time for high-level thinking, is a sufficiently significant improvement so far. On the other hand, engineering practices such as lab automation and manufacturing are in high demand for preciseness. This leads to the requirement of manual certification. Domain experts handle subtle cases through their tacit domain-specific knowledge and are responsible for their decisions~\citep{wang2023scientific}. According to these considerations and the standard operating processes of experimental sciences, we choose to certify the designed protocols by domain experts.

Our current choice is a compromise on the limitation of techniques and the demand for preciseness. In future work, we can conduct investigations on how to build digital twins of self-driving laboratories. Such digital twins support prediction, explanation, and counterfactual analysis of unseen behaviors of the experiments, which may facilitate machine-based protocol certification. Grounding these blue-sky thoughts necessitates addressing the challenging problems regarding the decision of simulation granularity, the implementation of data-efficient simulation model construction, and the injection of tacit domain-specific knowledge. In summary, the exploration of generated-protocol-certification by machines represents a promising avenue for future research.

\subsection{Limitations of automatic protocol certification}

\acp{llm} can be much too uncontrollable for engineering practices such as lab automation, which may lead to unpredictable dangerous situations~\citep{wang2023scientific}. There comes a dilemma---we try to exploit the capability of reasoning over knowledge of \acp{llm}, while we try to alleviate the drawbacks brought up by the uncontrollable nature of \acp{llm}. Our proposed representation is dedicated to resolving the dilemma. The representations not only elicit \acp{llm}' potential on protocol design through structural knowledge representation, but also serve as a guardrail for \acp{llm}. Since the generated protocols are represented as corresponding DSL programs, the permissible output space is much more confined compared with that of pure \acp{llm}, serving as constraints upon the \ac{llm}-generated protocols. Thanks to the verification mechanisms provided by \acp{dsl}, the correctness of the generated protocols can be checked to some extent. Therefore, by equipping \acp{llm} with an auxiliary constraint layer, we may approach a balance between knowledge utilization and preciseness.

However, the current verification on the level of \ac{dsl} programs is far from sufficient for serving as a certification. Certification is a serious process, where any possibilities of reporting false positive cases are required to be eliminated. Some cases can be highly long-tailed distributed, which may not be detected by data-driven and knowledge-driven machine certifiers. In this context, human domain experts are responsible for coming up with these potential risks through their experiences and tacit knowledge. Therefore, we are not likely to move human experts out of the loop, except that we can efficiently build up appropriate digital twins for self-driving laboratories. In current practices, the automation of protocol design puts human experts into a larger loop without focusing on the low-level details of experiments. As a result, they are allowed more time for high-level thoughts on things like values, which are not likely to be alternated by machines. In summary, it is neither practical nor necessary to totally move human experts out of the loop of automatic scientific discovery. The investigation of human-machine coordination in protocol certification represents a promising avenue for future research.

\subsection{Rationale for the reagent consumption model}

We treat the instantiation and the consumption of reagents a \emph{one-time deal} without considering the exact volume of consumption and the corresponding remainder. The rationale for such design choice comes from both the current \ac{sop} of experimental sciences and the properties of self-driving laboratories~\citep{bartley2023building}.

In the current \ac{sop} for manually conducted experiments, experimenters are required to use prefabricated sets of reagents. Similarly, experimenters use specific containers with predefined capacities to transfer intermediate products. Therefore, one pack of reagents or one container of intermediate products is only used once for an operation, without considering the remainder. This results in a more succinct representation where reagents are regarded as discrete elements rather than continuous volumes.

For self-driving laboratories, this is deliberately designed for efficient variable management following the corresponding principles in computer system design~\citep{abelson1996structure}. In computer systems, not removing used variables would cause out-of-memory errors, let alone in physical automation systems, where the physical memory slots are much harder than the virtual memory slots in computer systems to manage. Hence, we exploit this variable management mechanism to enhance the execution efficiency of self-driving laboratories.

\subsection{Relation to LLM reasoning}

We would like to clarify that our objective is not to alternate \ac{cot} reasoning. According to recent studies on the properties of \ac{cot}, \acp{llm} with \ac{cot} may generate coherent but unprofessional text in expertise-intensive application scenarios~\citep{xiao2023chain}. Therefore, our proposed representation serves as an auxiliary guardrail module for \acp{llm} with reasoning techniques such as \ac{cot}, enhancing \acp{llm}' reasoning capability from two aspects: (i) the representation constrain the scope of reasoning into a close set of entities, such as available operations, reagents, and devices commonly used in the domain; and (ii) the representation provides fine-grained injection of domain-specific knowledge for \acp{llm}, resulting in not only coherent but also expertise-compatible generated content.

\subsection{Applicability to domains beyond scientific experiment}

In theory, our framework can be applied to any field that requires adherence to specific protocols and has a need for automated execution. Let us consider an automated kitchen controlled by a computer as an example.

Assuming the automated kitchen’s computer is already programmed to prepare “braised pork ribs” and “steamed sea bass”:

\begin{lstlisting}[]
Braised Pork Ribs:

1.Select pork ribs as the main ingredient.
2.Heat a pan over high heat.
3.Add the ribs to the pan and fry for about 5 minutes until they are browned.
4.Add seasonings: soy sauce and sugar.
5.Reduce the heat to medium.
6.Simmer the ribs for 30 minutes until tender.
7.Serve hot.

START
SELECT ingredient: ribs
ACTION: fry, temperature: high, time: 5 min
ADD seasoning: soy sauce, sugar
ACTION: simmer, temperature: medium, time: 30 min
END

Steamed Sea Bass:

1.Select a whole sea bass as the main ingredient.
2.Prepare a steamer and heat it to high temperature.
3.Place the sea bass in the steamer.
4.Steam the fish for about 15 minutes until fully cooked.
5.Add seasonings: ginger slices and chopped scallions.
6.Serve immediately with the garnish.

START
SELECT ingredient: sea bass
ACTION: steam, temperature: high, time: 15 min
ADD seasoning: ginger, scallion
END
\end{lstlisting}

Next, we can derive the corresponding \ac{dsl}. For instance:

\begin{lstlisting}[]
{
  "cooking_methods": {
    "braise": {
      "steps": [
        {"type": "fry", "temperature": "high", "time": "5 min"},
        {"type": "simmer", "temperature": "medium", "time": "30 min"}
      ],
      "seasoning": ["soy sauce", "sugar"]
    },
    "steam": {
      "steps": [
        {"type": "steam", "temperature": "high", "time": "15 min"}
      ],
      "seasoning": ["ginger", "scallion"]
    }
  },
  "ingredients": {
    "ribs": {
      "category": "meat",
      "default_braise_time": "30 min"
    },
    "sea_bass": {
      "category": "fish",
      "default_braise_time": "20 min",
      "default_steam_time": "15 min"
    }
  }
}
\end{lstlisting}

Now, let us create a new recipe for Braised Sea Bass by combining the braising technique with sea bass as the main ingredient.

\begin{lstlisting}[]
START
SELECT ingredient: sea bass
ACTION: fry, temperature: high, time: 5 min
ADD seasoning: soy sauce, sugar
ACTION: simmer, temperature: medium, time: 20 min
END
\end{lstlisting}

\clearpage
\newpage
\section{Complete results}\label{sec:supp-result}

\subsection{Task-indexed complete results}\label{subsec:supp-result-task}

\input{tables/complete_results_task.sty}

\subsection{Domain-indexed complete results}\label{subsec:supp-result-domain}

\input{tables/complete_results_domain.sty}

\section{Ethics statement}\label{sec:supp-ethics}

\subsection{Human expert participants}

The testing set selection and groundtruth checking tasks conducted by human experts in this work has been approved by the \ac{irb} of Peking University. We have been committed to upholding the highest ethical standards in conducting this study and ensuring the protection of the rights and welfare of all participants. We paid the domain experts a wage of \$22.5/h for their work in this study. 

We have obtained informed consent from all human experts, including clear and comprehensive information about the purpose of the study, the procedures involved, the risks and benefits, and the right to withdraw at any time without penalty. Participants were also assured of the confidentiality of their information. Any personal data collected (including name, age, and gender) was handled in accordance with applicable laws and regulations. 

\subsection{Corpora collection}

We carefully ensure that all protocols included in our corpora strictly comply with open access policies under the Creative Commons license. This strategy guarantees adherence to copyright and intellectual property laws, thereby preventing any potential infringement or unauthorized use of protected materials. By exclusively employing resources that are freely accessible and legally distributable, we maintain the highest standards of ethical research conduct, promoting transparency and respect for the intellectual property rights of others. This commitment ensures that our work advances the frontiers of knowledge in a manner that is both legally sound and ethically responsible.

\section{Implementation details}\label{sec:supp-implement}

\subsection{Prior model of product flow-centric view}\label{subsec:supp-implement-prior-product}

\begin{lstlisting}[]
<ProductFlow> ::= <Pred> <FlowUnit> <Succ>

<Pred> ::= <Operation.UniqueName>

<Succ> ::= <Operation.UniqueName>

<FlowUnit> ::= <Component> <ComponentType> <RefName> <Vol> <Container> *<Cond>
<Component> ::= <STR>
<ComponentType> ::= Gas | Liquid | Solid | Semi-Solid | Mixture | ChemicalCompound | BiologicalMaterial | Reagent | PhysicalObject | File/Data | ... [Known component types]
<RefName> ::= <Component> <Index>
<UnitArgType> ::= MAT | PROD
<Vol> ::= <REAL> <MEAS>
<Container> ::= Tube | Flask | Pipette | ... [Known container types]
<Cond> ::= <ArgKey> <ArgValue>
<ArgKey> ::= Temperature | Pressure | Acidity | Lighting | ... [Known conditional keys]
<ArgValue> ::= <REAL> <MEAS>
\end{lstlisting}

\subsection{Prior model of operation-centric view}\label{subsec:supp-implement-prior-operation}

\begin{lstlisting}[]
<Operation> ::= <UniqueName> *<Pattern> 

<UniqueName> ::= <STR>

<Pattern> ::= <Precond> <Execution> <Postcond> *<Example>

<Precond> ::= <SlotArgNum> *<SlotArg>
<SlotArgNum> ::= <INT>
<SlotArg> ::= <ProductFlow.FlowUnit.ComponentType>

<Postcond> ::= <EmitArgNum> *<EmitArg>
<EmitArgNum> ::= <INT>
<EmitArg> ::= <ProductFlow.FlowUnit.ComponentType>

<Example> ::= <STR>

<Execution> ::= <DeviceType> <Capacity> *<Config>
<DeviceType> ::= Incubator | Autoclave | Centrifuge | ... [Known device types] 
<Capacity> ::= <REAL> <MEAS>
<Config> ::= <ArgKey> <ArgValue>
<ArgKey> ::= Duration | Pace | Power | Quantity | ... [Known device configuration items]
<ArgValue> ::= <REAL> <MEAS>
\end{lstlisting}

\subsection{Pre-processing of the protocols}\label{subsec:supp-implement-preprocess}

The protocol pre-processing steps begin by reading all JSON files of the protocols. Each protocol is then splitted sentence-by-sentence using \texttt{Spacy}\footnote{\url{https://spacy.io/api/sentencizer}}, with the constraint that every sentence is longer than ten characters. Due to the large volume of data, sentence splitting is handled in parallel. Afterwards, deeper sentence splitting is performed based on specific conditions for further refinement, such as the presence of \texttt{"and/then/and then"} followed by a verb\footnote{\url{https://spacy.io/api/matcher\#_title}}. We then parse sentences into root verbs and purpose clauses, which are identified using \texttt{token.dep\_ == "ROOT"} for root verbs and \texttt{prepositional/adverbial/modals} for purpose clauses. Lastly, we merge phrases based on punctuation, and their classification into valid sentences or decorative phrases depends on whether they contain a root verb or lack a purpose clause.

The first verb in each sentence is extracted as an opcode, again utilizing parallel processing for efficiency. Opcode frequency is filtered to exclude stopwords, which are recorded in a separate text file. Then we categorize these opcodes into high-level operation classes using a GPT model (gpt-4o mini), where each opcode is classified into categories like \texttt{Transfer Operations}, \texttt{Transformation Operations}, or \texttt{Data Operations}.

Once operation classification is complete, entity recognition is performed (also using gpt-4o mini) to identify entities like \texttt{devices}, \texttt{input\_flow\_units}, \texttt{output\_flow\_units}, and \texttt{total\_time}. Each flow unit is further categorized (also using gpt-4o mini)  with a high-level classification composed of a phase, \ie, \texttt{Gas}, \texttt{Liquid}, \texttt{Solid}, \etc; and a type, \ie, \texttt{Chemical Compound}, \texttt{Biological Material}, \etc. When both phase and type are successfully labeled, phase is preferred as the feature of the flow unit. If phase labeling fails, we use type the feature of the flow unit. If neither phase nor type is successfully labeled, the corresponding feature is set to \texttt{None}. Part of the rationale is that there are non-reagent components in the general sense, \ie, \texttt{data}, \texttt{files}, \texttt{obscure or undefined substances}, \etc. Therefore, we apply this strategy to maximize the possibility that there is a meaningful upper class labeling of the components without any redundancy.

Finally, we conduct a synonym merge process on the devices, which starts by using \texttt{transformers AutoTokenizer}\footnote{\url{https://huggingface.co/docs/transformers/v4.45.1/en/model_doc/auto\#transformers.AutoTokenizer}} to get an embedding for each device name. Afterwards, we use \texttt{sklearn}\footnote{\url{https://scikit-learn.org/stable/modules/generated/sklearn.metrics.pairwise.cosine_similarity.html\#cosine-similarity}} to identify potentially similar entity pairs by calculating the cosine similarity of the candidate entities, and then passing these entity pairs to the GPT model for synonym detection, thereby merging devices belonging to the same type. The reference names of these combined devices will be one of the features.

\subsection{Pure LLM-based designer}\label{subsec:supp-implement-baseline}

The pure \ac{llm}-based designer employs \ac{rag} to retrieve similar protocols from the corresponding corpora for representation, following the design choice of the baseline in~\citet{o2023bioplanner}. Specifically, in the \texttt{FB} approach, three similar protocols are first retrieved from the original protocol corpora using \ac{rag}, and then, along with the title and description of the target protocol, they are provided to the \ac{llm} to generate a \ac{nl} plan. The \ac{llm} subsequently translates the \ac{nl} plan into Python pseudocode. In the \texttt{IB} approach, three similar protocols’ instance actions (like Python pseudofunctions definitions) are first retrieved from the corpora, and after randomizing their order, they are provided to the \ac{llm} along with the title and description of the target protocol to generate a plan in the form of Python pseudocode.

\begin{lstlisting}
[Prompt for retrieving similar protocols from corpora]
You are an expert in biology and you are very familiar with the experiment protocols.
I would like to make a protocol for {title}. 
I will give you some related protocols in the database.
Could you find me the most three similar and relevant protocols for reference in the given range?

Please output id of your selected protocols, separating with a comma. Don't output any other information.
[Output format]
id_1,id_2,id_3

[Related protocols]
{context}

Answer:
\end{lstlisting}

\begin{lstlisting}
[Prompt for generating NL plan]
Your goal is to generate steps for a biology protocol.
These protocol steps must accurately describe a complete scientific protocol to obtain a result.
Steps of some similar protocols will be provided as a reference for you to generate the new one.
Output should only contain the steps without any other information.

Here is an example of how to generate steps for a biology protocol.

EXAMPLE:

{example protocol title}

Here are some extra details about the protocol:

{example protocol description}

example steps:

{example protocol steps}

YOUR TASK:
Generate steps for a protocol for {title}.

Here are some extra details about the protocol:

{details}

Here are some similar protocols' steps for reference:

{steps}

your steps:
\end{lstlisting}

\begin{lstlisting}
[Prompt for translating NL plan to pseudocode]
Your goal is to convert biology protocols into python pseudocode. 

EXAMPLE
Here is an example of how to convert a protocol for {example protocol title} into python pseudocode

{example protocol}

{example python pseudocode}

YOUR TASK:
Here is a biology protocol entitled '{title}' The protocol steps are as follows:

{protocol}

Please convert this protocol into python pseudocode.

python pseudocode:
\end{lstlisting}

\begin{lstlisting}
[Prompt for generating plan in pseudocode]
Your goal is to generate python pseudocode for biology protocols. 

Here is an example of how to generate pseudocode for a biology protocol.

EXAMPLE:

{example protocol title}

Here are some extra details about the protocol:

{example protocol description}

example pseudocode:

{example pseudocode}

YOUR TASK:
Generate pseudocode for a protocol for {title}.

Here are some extra details about the protocol:

{details}

You may only make use of the following python pseudocode functions:

{psuedofunctions}

your pseudocode:
\end{lstlisting}

\subsection{Internal designer}\label{subsec:supp-implement-internal}

The internal designer incorporates the specific representation as part of the prompt for an \ac{llm}, asking it to output the protocol while adhering to the given representation constraints, echoing the idea of~\citet{wang2023grammar}. Specifically, in \texttt{II}, the instance actions retrieved from the corpora via \ac{rag} and the pseudofunctions definitions of the target protocol are shuffled and then provided together to the \ac{llm}, constraining it to generate a plan in the form of Python pseudocode using the given pseudofunctions definitions. In \texttt{EI} and \texttt{EI+}, relevant \ac{dsl} instructions are selected from a domain-specific operation-centric view \ac{dsl} and product-flow-centric view \ac{dsl}, respectively. These instructions and the target protocol's title and description are provided to the \ac{llm}, prompting it to output the corresponding plan as instantiated DSL instructions.

\begin{lstlisting}
[Protocol for generating plan in DSL program using operation-centric view DSL]
Your goal is to generate plan in domain specific language (DSL) for biology protocols.
The DSL specifications related to the operations involved in the experiment are provided. The DSL specification of each operation consists of multiple patterns, each pattern is an operation execution paradigm.

Here is an example of how to generate plan in DSL for a biology protocol.

EXAMPLE:

{example protocol title}

Here are some extra details about the protocol:

{example protocol description}

example plan in DSL:

{example plan}

[Requirements]
1. Design the experiment with finer granularity, incorporating more steps to complete the experiment in a more rigorous, complex, and comprehensive manner.
2. There are some missing parameters in the DSL specification. You should generate each step of the DSL program as detailed as possible based on your understanding of the protocol plan.
3. In Precond and Postcond, use formal name of the component to represent the SlotArg and EmitArg of each step. The component name should clearly describe the content of the component.

YOUR TASK:
Generate plan in DSL for a protocol for {title}.

Here are some extra details about the protocol:

{details}

You can choose to instantiate the following DSL specification to construct the DSL program:

{DSL}

Your plan in DSL program:
\end{lstlisting}

\begin{lstlisting}
[Protocol for generating plan in DSL program using dual representation]
Your goal is to generate plan in domain specific language (DSL) for biology protocols.
Two perspectives of the DSL specification are provided: the specification for experimental operations and the specification for experimental products. 
The DSL specification of each operation or product consists of multiple patterns, each pattern is an operation execution paradigm or a product flow paradigm.
Output every operation of the plan in the form of an operation DSL program and every product of the plan in the form of a product DSL program.

Here is an example of how to generate plan in DSL for a biology protocol.

EXAMPLE:

{example protocol title}

Here are some extra details about the protocol:

{example protocol description}

example plan in DSL:

{example plan}

YOUR TASK:
Generate plan in DSL for a protocol for {title}.

Here are some extra details about the protocol:

{details}

You can choose to instantiate the following DSL specifications to construct the DSL program:

Operation-view DSL specification:
{Operation-DSL}

Product-view DSL specification:
{Product-DSL}

Your plan in DSL program:
\end{lstlisting}

\subsection{External designer}\label{subsec:supp-implement-external}

The external designer combines (i) deductive verification through \ac{dsl}; and (ii) self-improvement by the \ac{llm}~\citep{madaan2023self}. In \texttt{EE}, the external verifier is provided by the operation-centric view \ac{dsl} and performs checks on two main aspects: (i) whether the precondition of each operation is an intermediate product of a previous step rather than appearing from nowhere; and (ii) whether the postcondition of each operation is used in subsequent steps rather than being omitted. Similarly, in \texttt{EE+}, the external verifier is provided by the \ac{dsl} with a dual representation, focusing on cross-verifying the parallel dual tracks (the two perspectives of the DSL program). It checks whether the corresponding operation causes each \emph{status transition} of the product: (i) whether the product in each product-view program is the output of its preceding operation; and (ii) whether the product in each product-view program is the input for its succeeding operation. If a mismatch occurs, the verifier generates corresponding error messages, such as \textit{``Error: The product \{product\} required by operation \{operation\} at step \{i\} is not available from previous steps.''} These error messages are then fed into the feedback-refine loop as feedback for the \ac{llm} to revise the plan. The loop terminates when the program passes the verification or reaches the maximum number of iterations, and the best result is retained based on the verification information.

\begin{lstlisting}
[Prompt for refining the plan according to the feedback vertified by operation-centric view
DSL]
Your task is to improve a biology experimental protocol plan represented in domain-specific language (DSL) based on provided feedback.
The input plan in DSL consists of multiple DSL programs, each representing one step in the experimental protocol planning process, arranged in top-down order to indicate the execution sequence of operations. 
Each DSL program has the following format:
{
    "Operation": ,    // Operation verb
    "Precond": {      // Precondition for this step
        "SlotArgNum": ,   // Number of arguments for the precondition
        "SlotArg":        // Input product for this step
    },
    "Execution": {
        "DeviceType": ,   // Execution device for the operation
        "Config": {       // dict of execution arguments - values
            Argkey: Argvalues  
        }
    },
    "Postcond": {     // Postcondition for this step
        "EmitArgNum":,    // Number of arguments for the postcondition
        "EmitArg":        // Output product for this step
    }
}

The provided feedback indicates errors that occurred when compiling the DSL programs. You need to correct the program to ensure that the product is properly transferred between each step, i.e., the input product of each step must be the output from a previous step (except for the first step), and verify whether the output of each step is used as the input for subsequent steps (except for the final step).
If you believe the error in a particular step is due to the step preparing reagents rather than using a previous intermediate product, you can ignore this error.

Output your refined plan in DSL, returning a JSON block without any additional information or comments.

YOUR TASK:
Refine the plan in DSL for a protocol for {title}.

Here are some extra details about the protocol:

{details}

Refine the following plan:

{plan}

Here is the feedback of the plan:

{feedback}

Your refined plan in DSL:
\end{lstlisting}

\begin{lstlisting}
[Prompt for refining the plan according to the feedback vertified by DSL with dual representation]
Your task is to improve a Biology experimental protocol plan represented in domain-specific language (DSL) based on provided feedback.
The input plan in DSL consists of multiple DSL programs from two perspectives: operation-view and product-view. The DSL programs from these two perspectives alternate and constrain each other.

This is the format of a product-view DSL program:
// Each product view DSL program represents the state of the product at that moment.
{
    Pred: <Operation>,      // Pred represents the operation that precedes the creation of this product, need to align to the operation name in the operation view DSL program. If the product is in its initial state, return "".
    FlowUnit: {     // FlowUnit defines the properties of the product being processed.
        Component: ,    // Component represents the actual product or material being processed, need to be the formal name of the component.
        ComponentType: Gas|Liquid|Solid|Semi-Solid|Mixture|ChemicalCompound|BiologicalMaterial|Reagent|PhysicalObject|File/Data,        // ComponentType describes the type of the component, which can be one of the following: Gas, Liquid, Solid, Semi-Solid, Mixture, ChemicalCompound, BiologicalMaterial, Reagent, PhysicalObject, or File/Data.
        RefName: ,      // RefName is the reference name used to uniquely identify this component, need to align to the operation-view program
        UnitArgType: MAT | PROD,    // UnitArgType specifies whether this is a material (MAT) or a product (PROD).
        Vol: ,      // Vol represents the volume or quantity of the component.
        Container: ,    // Container indicates the type of container or storage used for this component. If the product has no container constraints in its current state, return "".
        Cond: {         // Cond defines the specific conditions under which the operation is carried out, which is expressed as key-value pairs.
            ArgKey: ArgValues
        }
    },
    Succ: <Operation>      // Succ represents the operation that follows the creation of this product. If the product is in its final state, return "".
}

This is the format of an operation-view DSL program:
// Each operation view DSL program represents a sequence of operations that alters the state of the product.
{
    Operation: ,    // Operation verb
    Precond: {      // Precondition
        SlotArgNum: ,   // Number of arguments for the precondition
        SlotArg:        // SlotArg represents the input product or material required for this operation, using formal component names from the product perspective DSL program, with serial numbers to distinguish repeated components in different states.
    },
    Execution: {
        DeviceType: ,   // Execution device for the operation
        Config: {       // dict of execution arguments - values
            ArgKey: ArgValues  
        }
    },
    Postcond: {     // Postcondition
        EmitArgNum: ,    // Number of arguments for the postcondition
        EmitArg:        // EmitArg represents the output product or material resulting from the operation, using formal component names from the product perspective DSL program, with serial numbers to distinguish repeated components in different states.
    }
}

The provided feedback indicates errors that occurred when compiling the DSL programs. You need to correct the program to ensure that the state changes of each product's RefName in the Product-view are caused by the corresponding operations in the Operation-view.
If you believe the error in a particular step is due to a mismatch in product names between the two perspectives rather than an actual error, you can ignore this error.

Output your refined plan in DSL, returning a JSON block without any additional information or comments.

YOUR TASK:
Refine the plan in DSL for a protocol for {title}.

Here are some extra details about the protocol:

{details}

Refine the following plan:

{plan}

Here is the feedback of the plan:

{feedback}

Your refined plan in DSL:
\end{lstlisting}

\subsection{Computing load of the machine designers}\label{subsec:supp-implement-load}

For automated representation generation, we primarily used GPT-4o mini with OpenAI’s Batch API\footnote{\url{https://platform.openai.com/docs/guides/batch/batch-api}} for preprocessing, incurring a cost of approximately \$60 across four domains. The design of the \acp{dsl} was executed on a MacBook with an M2 chip, running 1,000 iterations to ensure convergence. This process required an average of 55 seconds per iteration for the operation-centric view \ac{dsl} and an average of 2 seconds per iteration for the product-centric view \ac{dsl}. For the machine designer, we primarily utilized GPT-4o mini combined with \ac{rag} for design, with a total cost of approximately \$10 (7 methods, 140 protocols). In summary, the overall computational load is relatively low, highlighting the accessibility of our machine designers when utilizing the proposed representations and the corresponding automatic representation generation modules.

\section{Data collection}\label{sec:supp-dataset}

\subsection{Corpora sources}\label{subsec:supp-dataset-corpus}

The corpora $\mathcal{C}$ for the automatic generation of representations (\cref{subsec:automate-problem}) and the corpora for selecting the testing set (\cref{subsec:result-task}) are both retrieved from open-sourced websites run by top-tier publishers, including Nature's Protocolexchange\footnote{\url{https://protocolexchange.researchsquare.com/}}, Cell's Star-protocols\footnote{\url{https://star-protocols.cell.com/}}, Bio-protocol\footnote{\url{https://bio-protocol.org/en}}, Wiley's Current Protocols\footnote{\url{https://currentprotocols.onlinelibrary.wiley.com/}}, and Jove\footnote{\url{https://www.jove.com/}}. These sources compile a dataset of 15,837 experimental protocols across four domains: Genetics (8794 protocols), Medical (7351), Ecology (812), and Bioengineering (3597), with minimal overlap between them. We aggregated the corpora and analyzed the themes of the protocols according to the first- and second-level labels attached to them. We adopt measures to ensure that $\mathcal{C}$ is mutually exclusive with the testing set.

Other domains, such as Physics and Chemistry, are also representative domains of experimental sciences, besides Biology, Medical, and Ecology. The preliminary factor that restricts our current scope is data accessibility. Due to the higher cost of accessing the corpora of protocols for conducting physics and chemistry experiments, for example, mining the protocol from the ``method'' section of relevant published papers, we leave the application to Physics and Chemistry for future work.

\subsection{Eliminating the risk of data leaking}\label{subsec:supp-data-leak}

\begin{wrapfigure}{r}{0.4\textwidth}
   \centering
   \includegraphics[width=\linewidth]{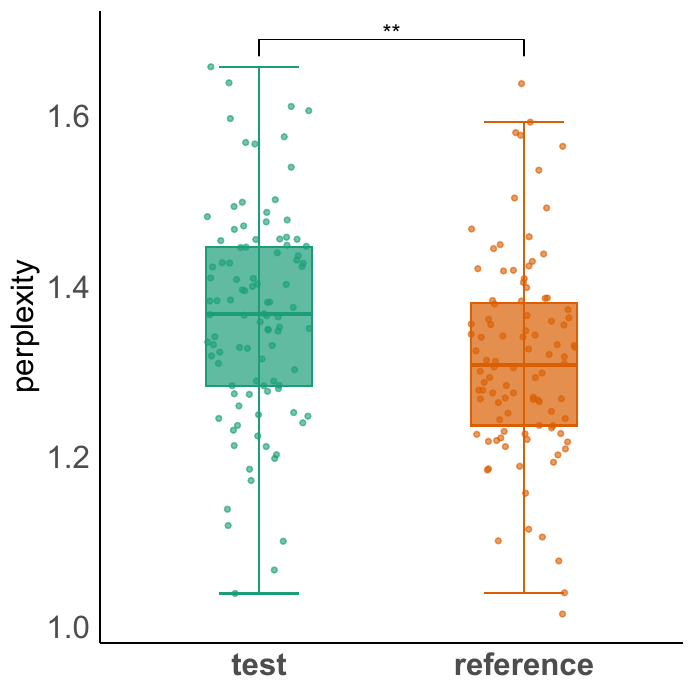}
   \caption{\textbf{Comparison between the perplexity of the test set and the reference set}}
   \vspace{-0.5\baselineskip}
   \label{fig:supp-perplexity}
\end{wrapfigure}

We employ the broadly accepted standard operating process to empirically verify that \acp{llm} have not memorized the data we use. We adopt the methodology outlined in Section 5.2 of \emph{Skywork}~\citep{wei2023skywork} and draw upon recent studies on detecting memorization in \acp{llm}~\citep{carlini2021extracting,carlini2022quantifying}. Specifically, we use gpt-4o mini to synthesize data resembling the style of steps from novel protocols, and then calculate the perplexity on the test set and reference set. Since the reference set is newly generated, we consider it clean, not belonging to any training set of any model.

We randomly sample 100 sequences each from the test set and the reference set of the novel protocols. Each sequence corresponds to a single procedural step described in \ac{nl}. We truncate the final 50 tokens of each sequence, retaining the prefixes. These prefixes are then used as prompts for the \ac{llm} to predict the next 50 tokens, for which we calculate the perplexity. If the perplexity of the test set is significantly lower than that of the reference set, the test set might have appeared in the model’s training phase.

The results indicate that the \ac{llm}'s average perplexity on the test set is significantly higher than that on the reference set ($t(198)=3.040, \mu_d<0, p<.05$; see \cref{fig:supp-perplexity}), suggesting that the \ac{llm} encounters greater uncertainty with the novel protocols in the test set. This finding implies that for a published, widely accepted, and standardized operating process, there is no evidence to suggest that the \ac{llm} has memorized the data.

\subsection{On the diversity of novel protocols}\label{subsec:supp-data-diversity}

\begin{figure}[ht]
    \centering
    \includegraphics[width=\linewidth]{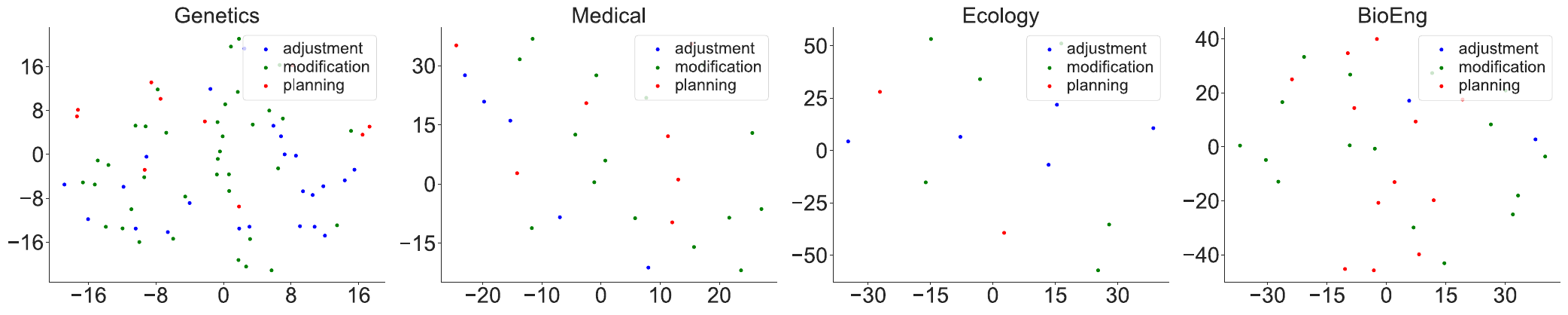}
    \caption{\textbf{Visualization of diversity between novel protocols}}
    \label{fig:supp-diversity}
\end{figure}

Assessing diversity among novel protocols is both informative and meaningful. To further support our analysis, we incorporate a t-SNE visualization of the experimental objectives (described in natural language) for the novel protocols we select, as shown at \cref{fig:supp-diversity}. The results demonstrate a well-dispersed distribution, indicating a sufficient level of diversity among the protocols.

\subsection{Showcases}\label{subsec:supp-dataset-showcase}

\input{cases/data_showcases.sty}

\section{Reproducibility}\label{sec:supp-reproduce}

The project page with supplementary files for reproducing the results of this paper will be available at \url{https://autodsl.org/procedure/papers/iclr25shi.html}.

\section{Limitations}\label{sec:supp-limit}

As a representation designed for a relatively new problem, the design and evaluation of the proposed framework come with limitations, leading to further investigations: 
\begin{itemize}[noitemsep,nolistsep,topsep=0pt,leftmargin=*]
    \item Overall, our method achieves promising results across the four domains. Specifically, it performs best in experimental design for Genetics, shows comparable effectiveness in Medical and Bioengineering, but is less effective in Ecology. Notably, the Genetics corpus is the largest among the four domains, while the Ecology corpus is significantly smaller than the others. These observations suggest a potential positive correlation between the size of the domain-specific corpus and the ``quality'' of the resulting \ac{dsl}. In other words, a larger corpus may lead to a ``better'' representation, thereby influencing the outcomes of protocol design. This hypothesis necessitates further investigation through rigorously designed experiments and carefully defined metrics for evaluating what constitutes a ``better'' representation. 
    \item We majorly consider the imperative programming \acp{dsl} as the implementation of representation in this work. This raises the question of whether incorporating objective-oriented programming paradigms could enhance the representation of complex entities within protocols, particularly the properties of reagents and intermediate products. If we are able to make the \acp{dsl} model the fine-grained reactions between different components and automate the design of those \acp{dsl} based on a broader source of data, such as the Wikipedia pages, we can ultimately manage to build up a symbolic digital twin for a domain-specific system, such as the cell cultivation environment. Such simulation systems may greatly benefit protocol design with their power of prediction, explanation, and counterfactual analysis.
    \item Can we explicitly extend our proposed representation to a hierarchical graph, thereby establishing the foundation for employing the advanced algorithms on graph routing and graph optimization? Results on the hierarchical graph can also serve as a external heuristic and constraint for \ac{llm}-based protocol designers. This hybrid approach may combine both the advantages of \acp{llm}, \ie, exploitation of background knowledge, and those of classical algorithms, \ie, white-boxed properties with high explainability. 
    \item Can we apply the representation and the automatic representation generator to other critical domains with a high demand for automating procedure design, such as designing product route sheets for advanced manufacturing?
\end{itemize}
With many questions unanswered, we hope to explore more on automated protocol design for self-driving laboratories and beyond.

\section{The automatically generated representations}\label{sec:supp-represent}

\subsection{Operation-centric view DSL}\label{subsec:supp-represent-operation}

\input{cases/operation_dsl.sty}

\subsection{Product-flow-centric view DSL}\label{subsec:supp-represent-product-flow}

\input{cases/product_dsl.sty}

\section{Case studies}\label{sec:supp-case}

\subsection{Case study: contributions of the building blocks}\label{subsec:supp-case-1}

\input{cases/building_blocks.sty}

\subsection{Case study: handling different task complexities}\label{subsec:supp-case-2}

\input{cases/scalability.sty}

\subsection{Case study: generality across domains}\label{subsec:supp-case-3}

\input{cases/generality.sty}

\end{document}